\documentclass[conference]{ieeeconf}
\IEEEoverridecommandlockouts    

\usepackage{multirow}
\usepackage{lipsum}
\usepackage{amsmath}
\usepackage{amssymb}
\usepackage[ruled,vlined]{algorithm2e}
\usepackage{graphicx}
\usepackage{subcaption}
\usepackage{multirow}
\usepackage{booktabs}
\usepackage{array}
\usepackage{url}
\usepackage{cite}
\graphicspath{{./images/}}
\usepackage[hidelinks]{hyperref}

\title{\LARGE \bf An Embedded Real-Time License Plate Recognition System for Complex Traffic Scenes}

\author{
    \parbox{\textwidth}{%
        \centering
        Anuki Pasqual$^{1}$, Dulan Lokugeegana$^{1}$, Manimohan Thiriloganathan$^{1}$, Nuthya Rathnayake$^{1}$, \\ Kithsiri Samarasinghe$^{1}$, and Udaya S. K. P. Miriya Thanthrige$^{1}$%
    }%
    \thanks{$^{1}$The authors are with the Dept. of Electronic and Telecommunication Engineering, University of Moratuwa, Sri Lanka
        {\tt\footnotesize \{pasqualac.20, lokugeeganadl.20, manimohant.20, rathnayakernp.20, kithsiris, sampathk\}@uom.lk}}%
    \thanks{© 2026 IEEE. Personal use of this material is permitted. Permission from IEEE must be obtained for all other uses, in any current or future media, including reprinting/republishing this material for advertising or promotional purposes, creating new collective works, for resale or redistribution to servers or lists, or reuse of any copyrighted component of this work in other works. Accepted at IEEE Intelligent Transportation Systems Conference (ITSC) 2026.}%
}
	\usepackage{draftwatermark}
	\SetWatermarkText{DRAFT}
	\SetWatermarkScale{1.0}  
	\SetWatermarkColor[gray]{0.88}  
\begin{document}
\bstctlcite{bstetal}

	\maketitle
	\thispagestyle{empty}
	\pagestyle{empty}
	
	\begin{abstract}
    Vehicle license plate recognition is an integral component of intelligent transportation systems. In this work, we present an embedded real-time license plate recognition system customized for developing countries. We address the challenge of handling complex, unstructured traffic scenes with diverse vehicle types while implementing the system on an embedded platform for low-cost deployment. Our method consists of license plate detection on a multi-vehicle image, followed by character recognition on the detected license plates. Both steps use lightweight convolutional neural networks to balance accuracy and efficiency. We also introduce the SL-LPR dataset of Sri Lankan road images, which contains a variety of vehicle types and traffic conditions typically seen in developing countries. On this dataset, the license plate detection and character recognition models achieved 93.6\% mAP and 87.88\% accuracy, respectively, and were competitive against larger models on several public datasets. To achieve real-time performance in a resource-constrained embedded environment, we applied low-bitwidth quantization using the Brevitas library and implemented FPGA acceleration for the models using the FINN framework. The end-to-end system can operate at 11.5~FPS when implemented on the Xilinx Kria KV260 platform. These results demonstrate that our system is effective for real-time license plate recognition on an embedded device, even in complex traffic scenarios. The SL-LPR dataset is available for research use at: \href{https://github.com/sl-lpr-uom/SL-LPR.git}{github.com/sl-lpr-uom/SL-LPR}.
    
	\end{abstract}

\setlength\textfloatsep{8pt}
\setlength\intextsep{8pt}
	
	\section{Introduction}
	\label{sec:introduction}
	Intelligent transportation systems require robust real-time license plate recognition (LPR) to enhance their functionality. This technology supports applications such as speed enforcement, road rule enforcement, traffic monitoring, and congestion charging systems. Implementing these systems in a developing country such as Sri Lanka can greatly improve road safety and transport efficiency in the country. However, there are unique considerations to be made in these countries compared to developed countries such as China and the United States, which have been studied extensively for existing LPR systems~\cite{xu2018towards, platedetection}. In developing countries, unstructured and heterogeneous traffic is more common. This leads to complex traffic scenes with no clear separation between lanes, and a larger number of small vehicles such as motorbikes. Furthermore, low-cost and low-power deployment is an important factor. Embedded systems with edge processing are useful for reducing cost and power, and for deployment in areas without stable network connections. A real-time LPR system that addresses the above challenges is essential for implementing intelligent transportation systems in developing countries.

{Several complexities arise when implementing a real-time system robust to these scenarios. Larger multi-vehicle images are needed to handle unstructured traffic scenes, which increases the computational complexity of the recognition method. In addition, the location of the license plate varies between different types of vehicles, requiring the method to consider a diverse range of vehicle types. Moreover, the limited computational resources of embedded systems make achieving real-time performance while addressing the aforementioned requirements a significant challenge.}

Prior studies have developed real-time embedded LPR systems primarily for controlled single-vehicle scenarios such as parking lot entrances and toll collection points, using classical image processing methods that are less adaptable to complex backgrounds~\cite{jeffrey_automatic_2013, rosli_real-time_2015, arth_real-time_2007}. A few multi-vehicle real-time implementations, which use deep learning methods, rely on embedded GPUs with higher power consumption~\cite{ammar_multi-stage_2023, lin_edge-ai-based_2022, park_all-one_2022}. While some studies address complex road scenarios with diverse vehicle types, they are limited to GPUs and high-performance CPUs, and the models used are too complex for real-time implementation on embedded devices~\cite{lin_efficient_2018, khan_performance_2021, laroca_robust_2018}. 

In this paper, we present a real-time embedded system for license plate recognition capable of handling complex road scenarios commonly found in developing countries. Our approach addresses the key challenges of non-lane-disciplined traffic, vehicle diversity, and limited computational resources, making it suitable for deployment in dynamic traffic environments. The contributions of our work are as follows:

\begin{itemize}
\item{Two lightweight convolutional neural network (CNN) models for license plate detection (LPD) and character recognition (LPCR) that accurately handle challenging road scenes using only 1.4 million total parameters.}

\item{The SL-LPR dataset, consisting of 2970 multi-lane road images and 3412 license plates captured from Sri Lankan roads, which includes diverse vehicle types and complex unstructured traffic scenarios.}

\item{A complete embedded implementation of the LPR system on the Kria KV260 platform with FPGA acceleration of the models, enabling real-time operation at 11.5 FPS with a maximum latency of 87 ms per frame.}

\end{itemize}

\begin{figure*}[htbp]
    \centering
    \includegraphics[width=1\textwidth]{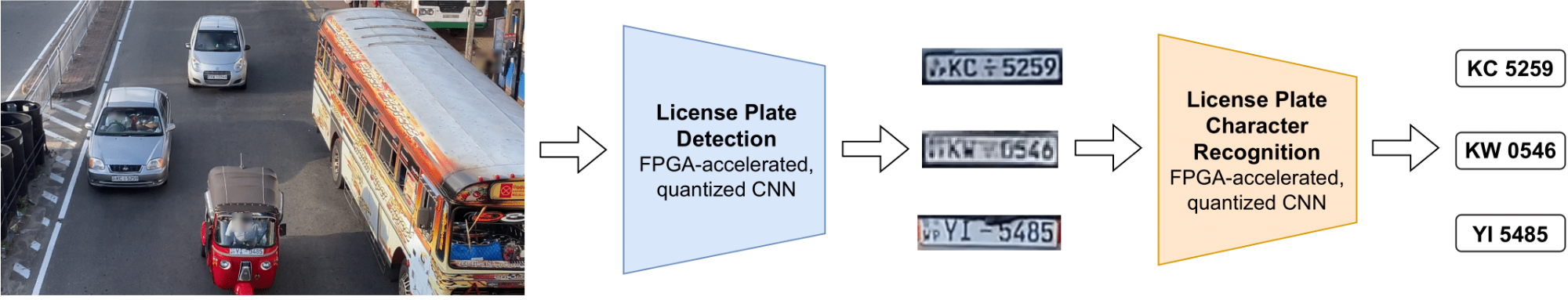}
    \caption{{High-level overview of the developed license plate recognition system.}}
    \label{fig:overallimg}
\end{figure*}

The developed models support easier implementation on embedded hardware due to their smaller size and quantized operations. The SL-LPR dataset and training techniques ensure the effectiveness of our system in complex road scenarios, and the FPGA-based implementation produces an efficient real-time license plate recognition system.

	\section{Related Work}
	\label{sec:relatedwork}
	This section reviews prior approaches for license plate detection and character recognition, focusing on works that address real-time embedded implementation, complex multi-vehicle scenarios, and the creation of suitable datasets.

Several approaches that target real-time edge-based computing utilize Jetson embedded GPU systems~\cite{ammar_multi-stage_2023, lin_edge-ai-based_2022, park_all-one_2022}. They use object detection methods based on You Only Look Once (YOLO) architectures for license plate detection and custom neural networks for character recognition. Embedded GPUs allow faster inference of larger networks, which is effective in multi-vehicle scenarios~\cite{park_all-one_2022}. However, they consume more power ( $\!>\!10\,\mathrm{W}$) and incur higher costs for large-scale deployment. Other real-time implementations employ FPGA~\cite{jeffrey_automatic_2013, rosli_real-time_2015} and DSP~\cite{jeffrey_automatic_2013, arth_real-time_2007} platforms for acceleration. These works apply classical image processing methods such as morphological operations, connected component analysis, and the AdaBoost algorithm. Although these methods are simpler to implement on custom accelerators, they have been evaluated only in single-vehicle scenarios in controlled environments such as parking lots. Multi-vehicle images introduce more background variability, which can lead to false detections when using less adaptable image processing algorithms~\cite{lin_efficient_2018}. Li et al.~\cite{li2024mixed} developed a transformer-based LPCR method effective in unconstrained settings. Although it uses FPGA acceleration, the 50~ms inference time per plate results in substantial latency for multi-vehicle scenes. Considering CPU-based embedded systems, Shashirangana et al.~\cite{shashirangana_license_2022} optimized a lightweight neural network for a Raspberry Pi through neural architecture search and achieved real-time performance with sufficient accuracy, yet this approach was also tested only on single-vehicle images. An ultra-low-power approach was developed by Lamberti et al.~\cite{lamberti_low-power_2021} by implementing CNNs for license plate recognition on a custom microcontroller platform, which only had a throughput of 1.09 FPS.

Many of the real-time embedded LPR systems use datasets with less complex scenes, containing only one or two vehicles per image~\cite{lin_edge-ai-based_2022}, or having homogeneous data with mostly car images~\cite{shashirangana_license_2022}. However, other studies explore LPR systems in more complex scenarios. Lin et al.~\cite{lin_efficient_2018} and Khan et al.~\cite{khan_performance_2021} implemented hierarchical approaches that perform vehicle detection first to avoid false positives in challenging multi-vehicle scenes. An improved YOLOv3 model was used by Liu et al.~\cite{liu_multi-object_2021} for multiple license plate detection in complex backgrounds with different license plate types. Usama et al.~\cite{usama_vehicle_2025} and Laroca et al.~\cite{laroca_robust_2018} created novel datasets with diverse vehicle types, including motorbikes, buses, and decorated trucks, to improve accuracy in heterogeneous traffic. Their works employed YOLO-based object detection models for license plate detection and character segmentation tasks.

The solutions that address complex road scenes have typically used neural network-based methods for license plate recognition, with their real-time implementations limited to GPUs. These networks are larger ($\sim$1~billion operations) and may not offer real-time performance on an embedded system. However, lightweight CNNs developed for character recognition, such as LPRNet~\cite{zherzdev2018lprnet} and fast-plate-ocr~\cite{fastplateocr}, show fast inference times on general-purpose CPUs and have been adapted into embedded systems~\cite{lamberti_low-power_2021}. Similarly, simplified YOLO-based object detection models developed for other real-time applications, such as face detection~\cite{gunay2022lpyolo}, can be adapted for license plate detection. 

To train and evaluate LPR methods for heterogeneous road conditions, a dataset reflecting such scenarios is required. Table~\ref{tab:dataset} shows a comparison between several publicly available LPR datasets, considering aspects relevant to challenging traffic scenes. Although CCPD~\cite{xu2018towards} is the largest public LPR dataset, the data consists completely of parked cars and does not represent traffic conditions seen in developing countries. Similarly, PKU~\cite{chinesedataset} and AOLP~\cite{hsu2012application} datasets predominantly contain car images with only 1-2 vehicles per image. The UFPR-ALPR dataset~\cite{laroca_robust_2018} shows greater variety in vehicle types and license plate positions; however, it only contains 150 unique vehicles, which may restrict generalization during training. The SL-LPR dataset addresses the limitations of these datasets by providing a high variety in vehicle types and license plate position relative to the camera, as well as having up to 8 license plates in a single scene. Hence, it represents the unstructured nature of traffic scenes in a developing country better than existing datasets.

The reviewed literature shows that the challenges of real-time embedded LPR systems and complex road scenes have not been addressed in combination by prior work. 

\newcolumntype{C}[1]{>{\centering\arraybackslash}m{#1}}
\newcolumntype{R}[1]{>{\raggedleft\arraybackslash}p{#1}}

\begin{table*}[htbp]
\centering
\caption{Comparison between public LPR datasets and our dataset, considering aspects relevant to complex traffic scenes.}
\label{tab:dataset}
\begin{tabular}{p{2.3cm}C{1.3cm}C{1.0cm}C{1.0cm}C{1.1cm}C{1.3cm}C{1.3cm}C{1.6cm}C{1.7cm}C{0.8cm}}
\toprule
\textbf{Dataset} & \textbf{Country} & \textbf{\# LPD images} & \textbf{\# LPCR images} & \textbf{\# unique LPs} & \textbf{Avg. LPs per image} & \textbf{Max LPs per image} & \textbf{Vehicle type variation} & \textbf{LP position/ size variation} & \textbf{Scene}\\
\midrule
CCPD~\cite{xu2018towards} & China & \textbf{290k} & \textbf{290k} & \textbf{270k} & 1.0 & 1 & Cars only & \textbf{High} & Parked\\
PKU dataset~\cite{chinesedataset} & China & 3924 & 4294 & 3828 & 1.1 & 2 & Cars, trucks & Medium & Road\\
UFPR-ALPR~\cite{laroca_robust_2018} & Brazil & 4500 & 4500 & 150 & 1.0 & 1 & \textbf{5+ types} & \textbf{High} & Road\\
AOLP dataset~\cite{hsu2012application} & Taiwan & 2049 & 2066 & 1291 & 1.0 & 2 & Cars only & Low & Road\\
\midrule
\textbf{SL-LPR (Ours)} & Sri Lanka & 2970 & 3412 & 1963 & \textbf{1.9} & \textbf{8} & \textbf{5+ types} & \textbf{High} & Road\\
\bottomrule
\end{tabular}
\end{table*}

	\section{Methodology}
	\label{sec:methodology}

    {In this section, we present our methodology for an efficient real-time license plate recognition system. Figure~\ref{fig:overallimg} presents an overview of our approach, which consists of performing license plate detection on the multi-lane image, cropping the detected plates, and recognizing the characters of each license plate. The detection and recognition steps are implemented as quantized CNN models and deployed on an FPGA for faster inference. We also describe our procedure for creating the SL-LPR dataset, as well as the embedded hardware implementation of the complete LPR system.}

\subsection{Dataset Creation}

 As existing datasets did not capture the variety of vehicle types and complex non-lane-disciplined road scenes, we created the SL-LPR dataset of Sri Lankan vehicle and license plate images. We captured images from overhead bridges on an urban road and an expressway, ensuring that the dataset covers all possible vehicle types and traffic levels. Data was captured in both daytime and low-light (dusk) conditions.

The camera parameters used for capturing images and the resulting specifications are shown in Table~\ref{tab:camera}. The resolution was set based on the road width such that the license plates in a multi-lane image are readable. The frame rate and shutter speed were set to ensure that vehicles traveling at a maximum of 250 km/h can be captured in at least two frames without motion blurring of the license plate. 

\begin{table}[ht]
    \centering
    \caption{Camera parameters and resulting specifications.}
    \label{tab:camera}
    \begin{tabular}{ll}
        \toprule
        \textbf{Frame rate}   & $10$ FPS  \\
        \textbf{Resolution}  & $1920 \times 1080$ \\
        \textbf{Shutter speed}    & $1/1200$ s   \\
        \midrule
        \textbf{Maximum vehicle speed} & $250$ km/h   \\
        \textbf{Road area captured} & $6$ m width, $12$ m length \\
        \bottomrule
    \end{tabular}
\end{table}

\begin{figure}[ht]
    \centering
    \includegraphics[width=0.99\linewidth]{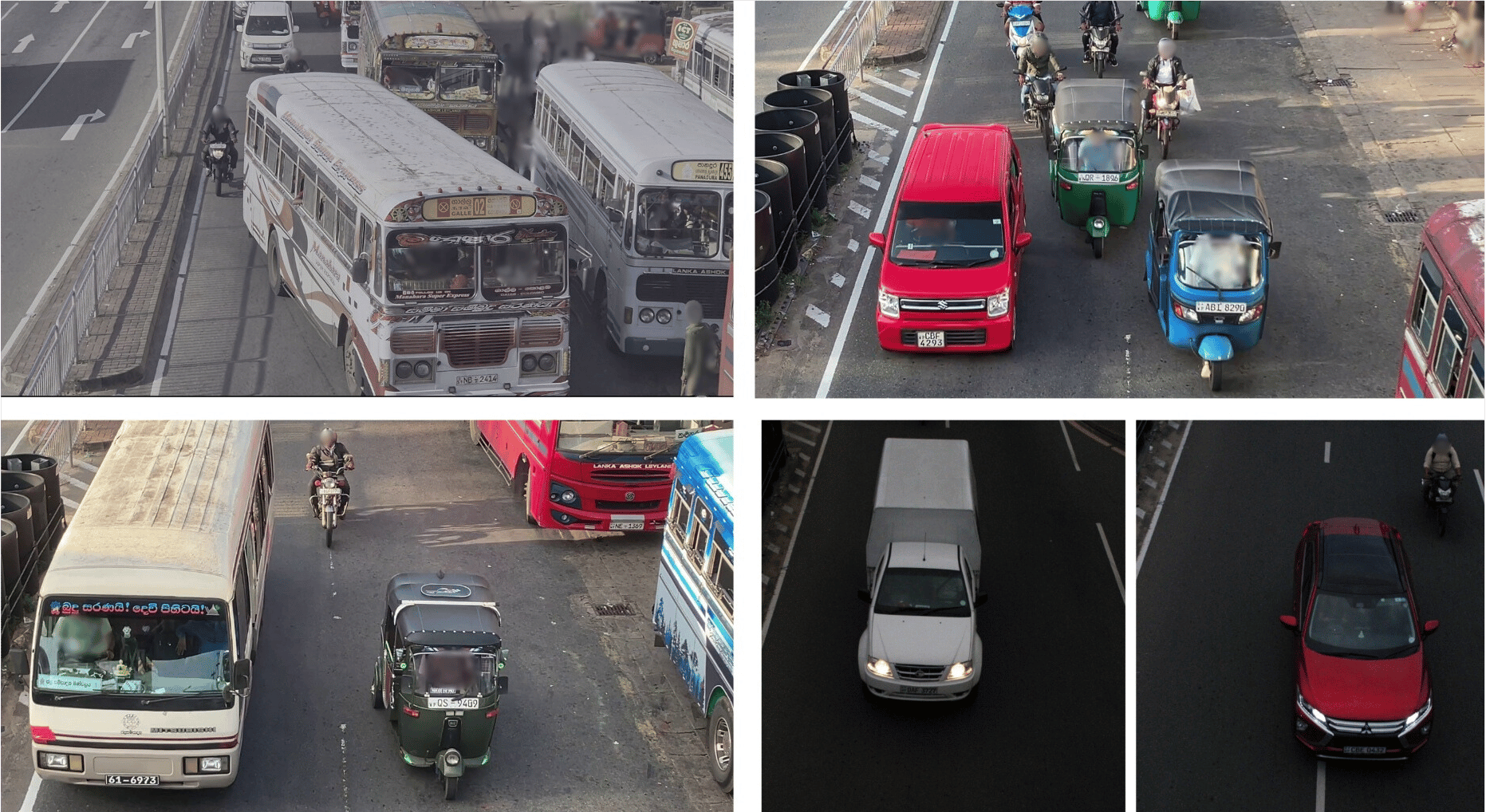}
    \caption{{Multi-lane images from the SL-LPR dataset captured on Sri Lankan roads, covering various vehicle types and traffic levels. People in the images are blurred for privacy.}}
    \label{fig:roadimages}
\end{figure}

{To make the dataset robust to any scenario, we captured scenes containing different types of vehicles commonly seen in Sri Lanka, including cars, motorbikes, buses, trucks, and three-wheelers (auto rickshaws). The images were taken under different lighting conditions and varying traffic levels to produce complex scenes. Figure~\ref{fig:roadimages} shows several images from the dataset that highlight these features. After filtering close duplicates, the dataset contained 2970 images with annotated bounding box labels. The images were divided as 80:18:2 for training, validation, and testing. 100 images captured on a different day were also added to the test set.}

\begin{figure}[htbp]
    \centering
    \includegraphics[width=0.99\linewidth]{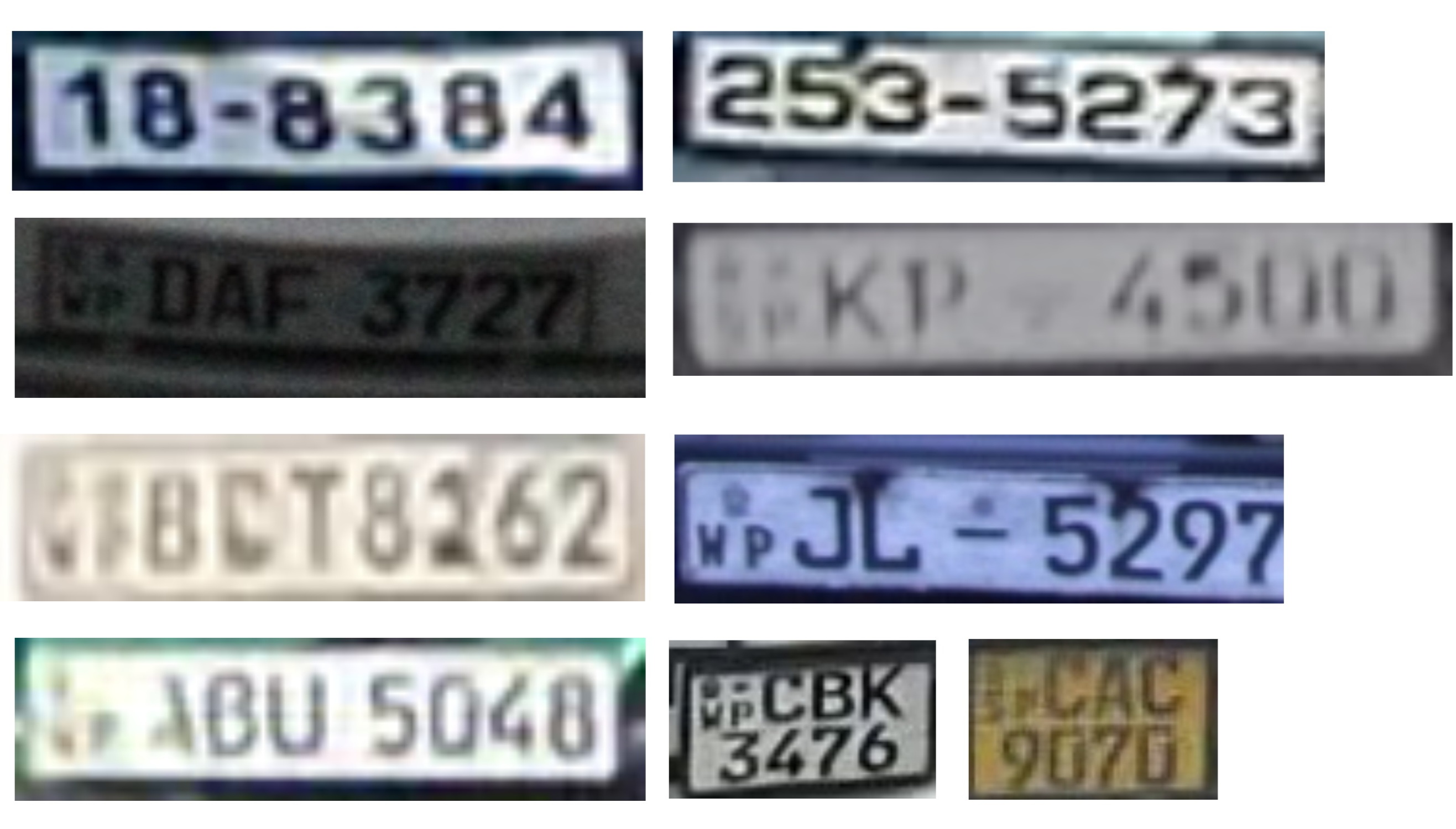}
    \caption{{License plate images from the SL-LPR dataset, covering different number formats and different clarity levels. }}
    \label{fig:plates}
\end{figure}

{
We used the labeled bounding boxes to extract only the license plates and created another dataset to be used for character recognition. We labeled the images manually, as general-purpose OCR libraries were only able to recognize 19\% of the plates correctly. With multiple occurrences of the same license plate, we kept at most 3 images with different clarity levels to reduce bias. After removing unreadable and repeated images, the dataset contained 3412 license plates. We divided the data as 80:20 for training and validation, and used a separate set of 165 images captured on a different day for testing. Figure~\ref{fig:plates} shows several images from the dataset with different number formats and clarity levels.}

\subsection{License Plate Detection}

{
License plate detection methods based on models such as YOLO provide better accuracy but require more processing, in the order of 1 billion floating-point operations~\cite{8492857}. Considering our requirement for real-time processing in an embedded device, we adapted the LPYOLO (low-precision YOLO) face detection architecture developed by Gunay et al.~\cite{gunay2022lpyolo} for license plate detection. The model was implemented using the Brevitas library~\cite{brevitas} to employ low-bitwidth quantized weights and activations, significantly reducing the model storage size and enabling easier hardware implementation.
}

\begin{figure}[h!]
    \centering
    \includegraphics[width=0.99\linewidth]{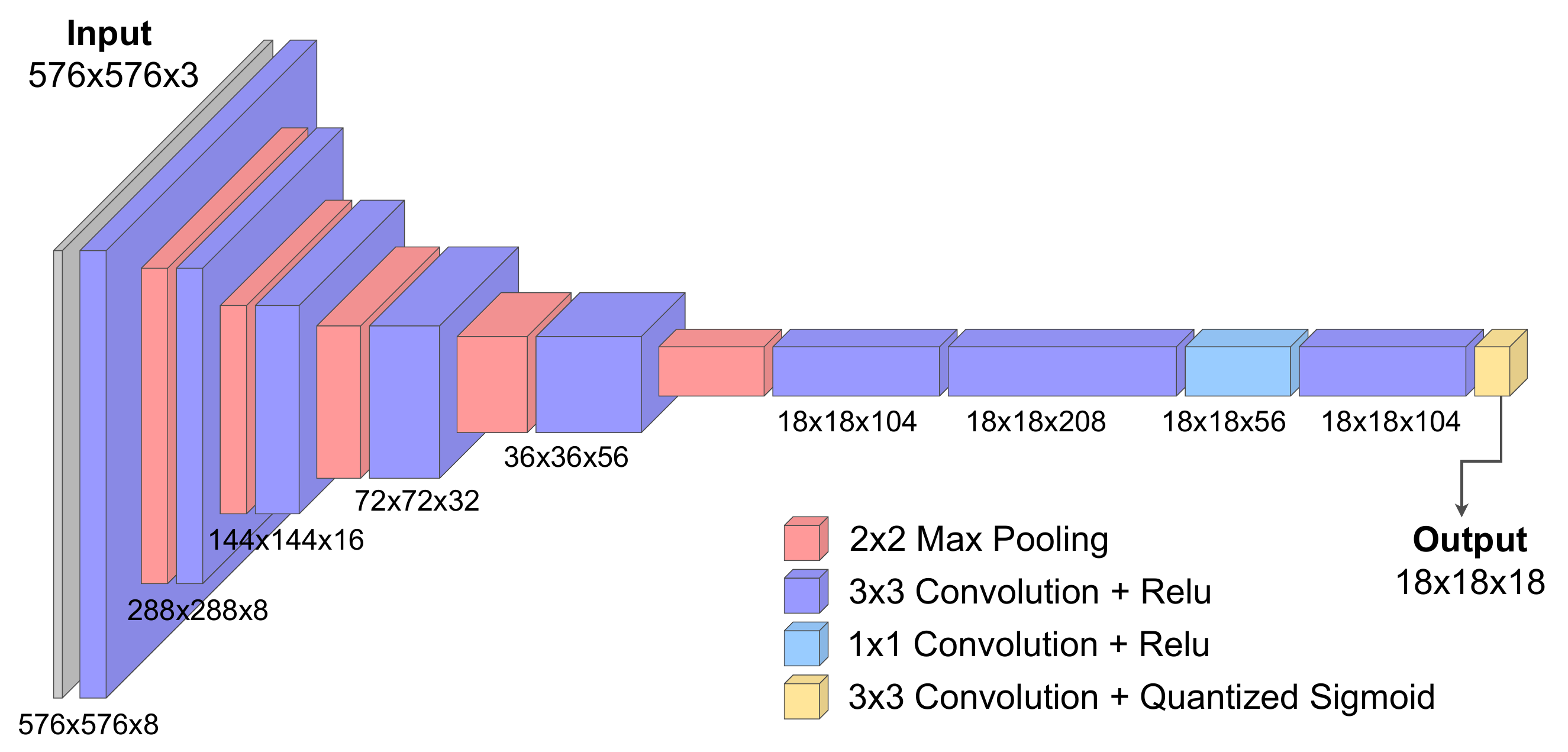} 
    \caption{{Architecture of the license plate detection model.}}
    \label{fig:lpdmodel}
\end{figure}

{
Figure~\ref{fig:lpdmodel} shows the architecture of the LPYOLO model~\cite{gunay2022lpyolo} that we adapted for license plate detection. Although the model is based on the YOLO model family, it has been simplified to use only 0.363 billion operations. Furthermore, all weights and intermediate outputs are quantized to 4 bits. We used a 576$\times$576 input image size as opposed to the 416$\times$416 input in the original model. This is because the dimensions of far-away license plates in the multi-lane image can become impossible to detect with a smaller image size. While 640$\times$640 is a common choice in YOLO models, we achieved sufficient accuracy with 576$\times$576, and increasing the resolution further will add unnecessary latency.  In the output, the image is divided into 18$\times$18 grid cells, each having predictions relative to 3 different anchor boxes. Each prediction consists of the center x and y coordinates, box width and height, class information, and confidence level. This structure results in an 18$\times$18$\times$18 output size. After filtering the predictions by confidence level, non-maximum suppression (NMS) is used to remove overlapping predictions and obtain the final result. 
}

{
The anchor boxes of the model were fine-tuned for our purpose. The LPYOLO model used anchors that were computed to fit a dataset that contains many generic object types~\cite{gunay2022lpyolo}. However, the aspect ratios and sizes of these boxes were significantly different from the typical size and shape of a license plate, which reduced the accuracy of predicted bounding boxes. Hence, we computed 3 new anchors by performing K-means clustering on the labels of the multi-lane license plate dataset. These anchors have aspect ratios similar to license plates, and the sizes correspond to small, medium, and large license plates found in multi-lane images. 
}


The last layer of the LPYOLO model uses the sigmoid activation function during training, but replaces it with a piecewise tanh function (HardTanh) during inference to avoid expensive computations~\cite{gunay2022lpyolo}. However, we found that the pixel error caused by this replacement is more noticeable in our application due to the smaller size of the license plates. Therefore, instead of HardTanh, we applied 8-bit quantization after limiting values to the range $[-3.5, 3.5]$. Then, sigmoid is applied to the quantized values, allowing us to precompute the 256 required sigmoid values. The limit value 3.5 was selected since $\mathrm{sigmoid}(3.5) = 0.97\simeq1$ reduces the error for saturated values while keeping the quantization error relatively low. The comparison in Table~\ref{tab:lpd_activation} shows that using HardTanh causes a large drop in accuracy, while our approach yields an accuracy level close to using an unquantized sigmoid function.

\begin{table}[ht]
\centering
\caption{Comparison of efficient final layer activation functions that can be used instead of sigmoid during inference, showing maximum pixel error of the prediction relative to the sigmoid baseline and the resulting average precision (AP).}
\label{tab:lpd_activation}
\begin{tabular}{@{}lcc@{}}  
\toprule
\textbf{Activation function} & \textbf{Max pixel error}  & \textbf{AP [IoU $\geq$ 0.5]} \\
\midrule
Unquantized sigmoid                  & Baseline   & 94\%     \\
HardTanh~\cite{gunay2022lpyolo}       & 15     & 52\%     \\
\textbf{8-bit quantization $\rightarrow$ sigmoid} & \textbf{3}  & \textbf{93.6}\%    \\
\bottomrule
\end{tabular}
\end{table}

{We pre-trained the model on the PKU dataset~\cite{chinesedataset} to enable reliable convergence before fine-tuning on our more complex dataset. During training, we applied the data augmentations of translation, horizontal flip, scaling, and HSV variations to assist the model in learning to detect license plates in various sizes, locations, and environments. }

\subsection{License Plate Character Recognition}

{
From reviewing prior work, we identified that character recognition methods based on CNNs can be both accurate and efficient. General-purpose OCR libraries such as EasyOCR~\cite{easyocrlib} use large models with higher complexity. Since we are targeting a real-time embedded implementation, our model is an adaptation of the simpler CNN architecture from the fast-plate-ocr library~\cite{fastplateocr}. We found that this method had better accuracy and better generalization compared to classical image processing methods, while still being lightweight.}

\begin{figure}[h!]
    \centering
    \includegraphics[width=0.99\linewidth]{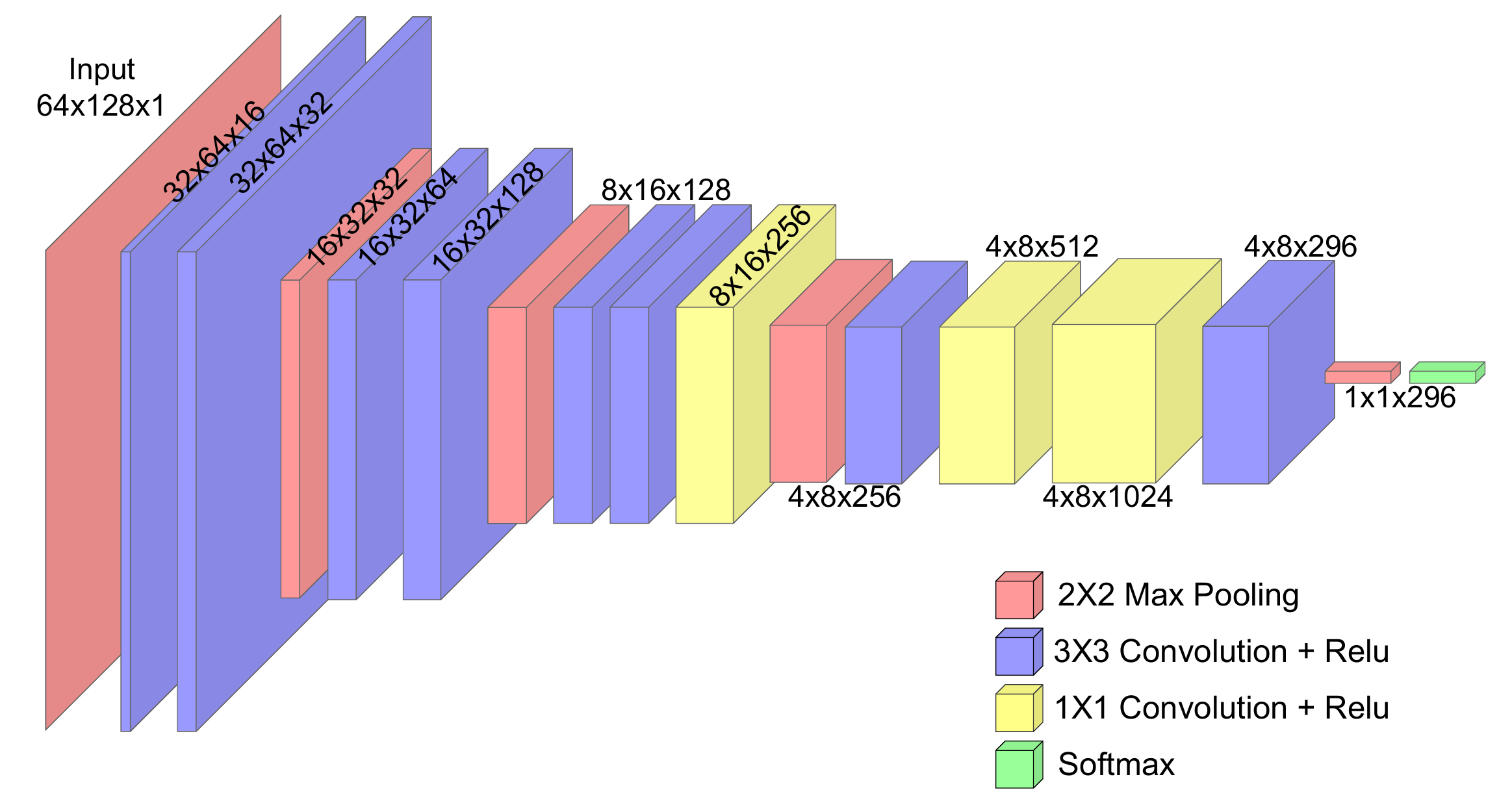}
    \caption{{Architecture of the character recognition model.}}
    \label{fig:ocr_model}
\end{figure}

{Figure~\ref{fig:ocr_model} shows the architecture of our character recognition model, which was created by optimizing the fast-plate-ocr~\cite{fastplateocr} architecture. To enable a simpler hardware implementation, we replaced the set of parallel fully connected layers used in the original model by a convolution layer, and used normal convolution instead of strided convolution. Furthermore, we reduced the number of model parameters by removing the first few $1 \times 1$ convolution layers and keeping only one layer with 1024 filters. Moreover, we observed that there are many filters with completely zero weights after training. Hence, we used model pruning to reduce the model size further. Table~\ref{tab:model-comparison} compares the number of parameters and multiplications of the original model (quantized) with our model after simplification and pruning, where the model size has been reduced by 91\%.

\begin{table}[ht]
    \centering
    \caption{Comparison between the original fast-plate-ocr model and our simplified and pruned LPCR models.}
    \label{tab:model-comparison}
    \begin{tabular}{lccc}
        \toprule
        \textbf{Model} & \textbf{Parameters} & \textbf{Multiplications} & \textbf{Accuracy} \\
        \midrule
        Original~\cite{fastplateocr}   & 11.47\,M   & 527\,M   & 89.70\% \\
        Simplified  & 4.25\,M & 258\,M   & 88.48\% \\
        After pruning    & \textbf{1.02\,M} & \textbf{132\,M}  & \textbf{87.88\%} \\
        \bottomrule
    \end{tabular}
\end{table}

The number of output channels was determined by the length of the license plate format and the number of character possibilities for each position. To account for variable license plate number length, spaces are used as padding characters in shorter plate formats. For each channel, global max pooling followed by the softmax activation function is used to determine the prediction for each character.  


We implemented the model using the Brevitas library~\cite{brevitas} to incorporate quantized weights and activations for all layers. This reduces the size of the model and the computational complexity. As Brevitas allows custom, arbitrary bit widths for weights, we used 4 bits in the first 3 layers, 2 bits in the next 7 layers, and 1 bit in the final layer. These bit widths were experimentally selected to ensure that accuracy is maintained. We used 4-bit quantization for all activations except the final layer, where we trained the model with 8-bit quantization to improve learning but used 4-bit quantization during inference. An input image size of 64 $\times$ 128 was used since Brevitas requires the dimensions to be divisible by the pooling stride.

To enable accurate predictions in low-light conditions, we used an intensity transformation after computing the average pixel intensity. This transformation will improve contrast in dark images while minimally affecting bright images, as shown in Figure~\ref{fig:intensity-both}. Moreover, we used data augmentations such as random rotation, blurring, and brightness variation to make our model robust to changes in lighting and orientation. 

\begin{figure}[htbp]
  \centering
  \subfloat[Low light\label{fig:intensity-low}]
           {\includegraphics[width=0.45\linewidth]{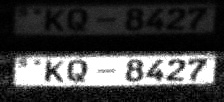}}
  \hspace{10pt}
  \subfloat[Normal\label{fig:intensity-normal}]
           {\includegraphics[width=0.4\linewidth]{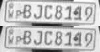}}
  \caption{Intensity transformation for two lighting conditions.}
  \label{fig:intensity-both}
\end{figure}


\subsection{Hardware Implementation}

By using the Brevitas library for both license plate detection and character recognition, the trained models can then be used with the Xilinx FINN framework~\cite{finn} to implement a custom accelerator for model inference on an FPGA. This utilizes parallel computations to significantly improve the latency and throughput compared to a CPU implementation when performing inferences in real time. We used this framework to implement the LPR pipeline on the Xilinx Kria KV260 module, which contains a Zynq Ultrascale+ MPSoC integrating ARM processor cores with a programmable FPGA fabric.

\begin{figure}[t]
    \centering
    \includegraphics[width=0.96\linewidth]{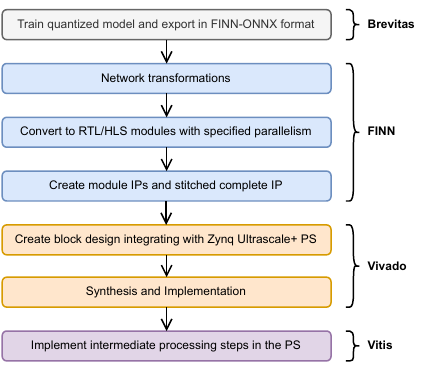}
    \caption{{Design flow of the FPGA implementation process.}}
    \label{fig:design_flow}
\end{figure}


Figure~\ref{fig:design_flow} shows the steps we followed to implement the license plate detection and character recognition models on the FPGA. First, the quantized model is trained using Brevitas as described in the above sections. The model is then converted to the FINN-ONNX format and given to the FINN framework for compilation into an FPGA design. In the FINN compilation process, first, network transformations are applied to the ONNX graph to simplify it into a set of hardware-implementable blocks. Then, these blocks are converted to RTL or HLS modules with the level of parallelism specified through a configuration file. Finally, module IPs are created and stitched together to make the complete IP for the model. Afterwards, Vivado is used to create a block design integrating both model IPs with the Zynq Ultrascale+ processing system (PS) through Direct Memory Access (DMA) modules. This design is synthesized and implemented for the FPGA. Since the FPGA accelerator is driven by the processing system, the driver code for handling model inputs and outputs is implemented as a C program using Vitis. Intermediate processing steps that are not part of the model, such as resizing, output post-processing, non-maximum suppression, and license plate cropping, are also implemented in the C program.

Since our approach involved implementing separate accelerators for license plate detection and character recognition in the same FPGA, significant optimizations were made in the FINN compilation process to ensure that both designs could fit within the resource constraints of the selected FPGA without sacrificing throughput. This included optimizing implementation configurations and reducing model parameter storage. Within the available resource constraints, we increased the parallelization parameters of the FPGA design to maximize the throughput. The final FPGA resource utilization was 63\% of LUTs, 38\% of FFs, 99\% of BRAM, and 29\% of DSPs.

	\section{Results and Discussion}
	\label{sec:results}

{
This section discusses the performance of the implemented license plate recognition system in terms of accuracy and inference latency. The models were trained using a Quadro RTX~6000 GPU with 24 GB of memory, and implemented on the Xilinx Kria KV260. Accuracy was evaluated on the SL-LPR dataset and several publicly available datasets.
}

\subsection{License Plate Detection}

{The LPD model was pre-trained for 300 epochs on the PKU dataset~\cite{chinesedataset} and finetuned for up to 700 epochs on the dataset used for evaluation. The learning rate was ramped up to 0.0035 within 50 epochs and then decayed down to 0.0015. The evaluation metrics used are the recall rate and average precision (AP), with correct predictions defined as having Intersection over Union~(IoU)~$\geq$~0.5. On the SL-LPR dataset, our model achieved 93.6\% AP and 92.2\% recall.  }


\begin{table}[htbp]
\centering
\caption{Comparison of our LPD model with other implementations based on recall accuracy (\%) on several public datasets and latency as reported in the original work.}
\label{tab:lpd_performance}
\begin{tabular}{lC{0.6cm}C{0.6cm}C{0.6cm}C{0.6cm}C{1.4cm}}
\toprule
\textbf{Method} & \textbf{PKU} & \textbf{UFPR-ALPR} & \textbf{AOLP} & \textbf{SL-LPR} & \textbf{Latency (ms)} \\
\midrule
Yuan et al.~\cite{chinesedataset}        & 97.7  & -- & -- & -- & 36 [CPU] \\
Laroca et al.~\cite{laroca_robust_2018}  & --  & \textbf{98.3} & -- & -- & 4 [GPU] \\
Hsu et al.~\cite{hsu2012application}  & --  & -- & 94.0 & -- & 90 [CPU] \\
Xie et al.~\cite{xie2018new}  & 97.4  & -- & \textbf{99.5} & -- & 530 [CPU] \\
Khan et al.~\cite{khan_performance_2021} & \textbf{99.6} & -- & 99.4 & -- & 570 [N/A]\\
\midrule
\textbf{Ours}                            & 98.6 & 96.7 & \textbf{99.5} & \textbf{92.2} & \textbf{10 [FPGA]} \\
\bottomrule
\end{tabular}
\end{table}

Table~\ref{tab:lpd_performance} compares our model with other LPD implementations evaluated on the PKU~\cite{chinesedataset}, UFPR-ALPR~\cite{laroca_robust_2018}, and AOLP~\cite{hsu2012application} datasets. Our model achieves competitive performance compared to prior work in all cases. Considering inference latency, our embedded FPGA implementation is faster than other models implemented on general-purpose CPU systems. The accuracy of our model on the SL-LPR dataset is lower than on other datasets, which is primarily due to the higher proportion of small license plates and the larger number of vehicles per image found in our dataset. 

{
Figure~\ref{fig:testresults} shows the model predictions against the ground truth bounding boxes in several complex traffic images from the SL-LPR test set. Nearly all the license plates have been accurately detected, except for a few smaller plates that are further away from the camera. However, this is acceptable since in a practical application, this model will be used with a continuous stream of images, and these smaller plates can be detected in later frames when the vehicle gets closer to the camera. Moreover, in several plates that are further away, the predicted bounding box is slightly off from the ground truth, which can cause a letter to be cropped out. This is handled by increasing the size of the bounding box by 5\% before cropping the plate for the LPCR step.
}

\begin{figure}[htbp]
    \centering
    \includegraphics[width=0.99\linewidth]{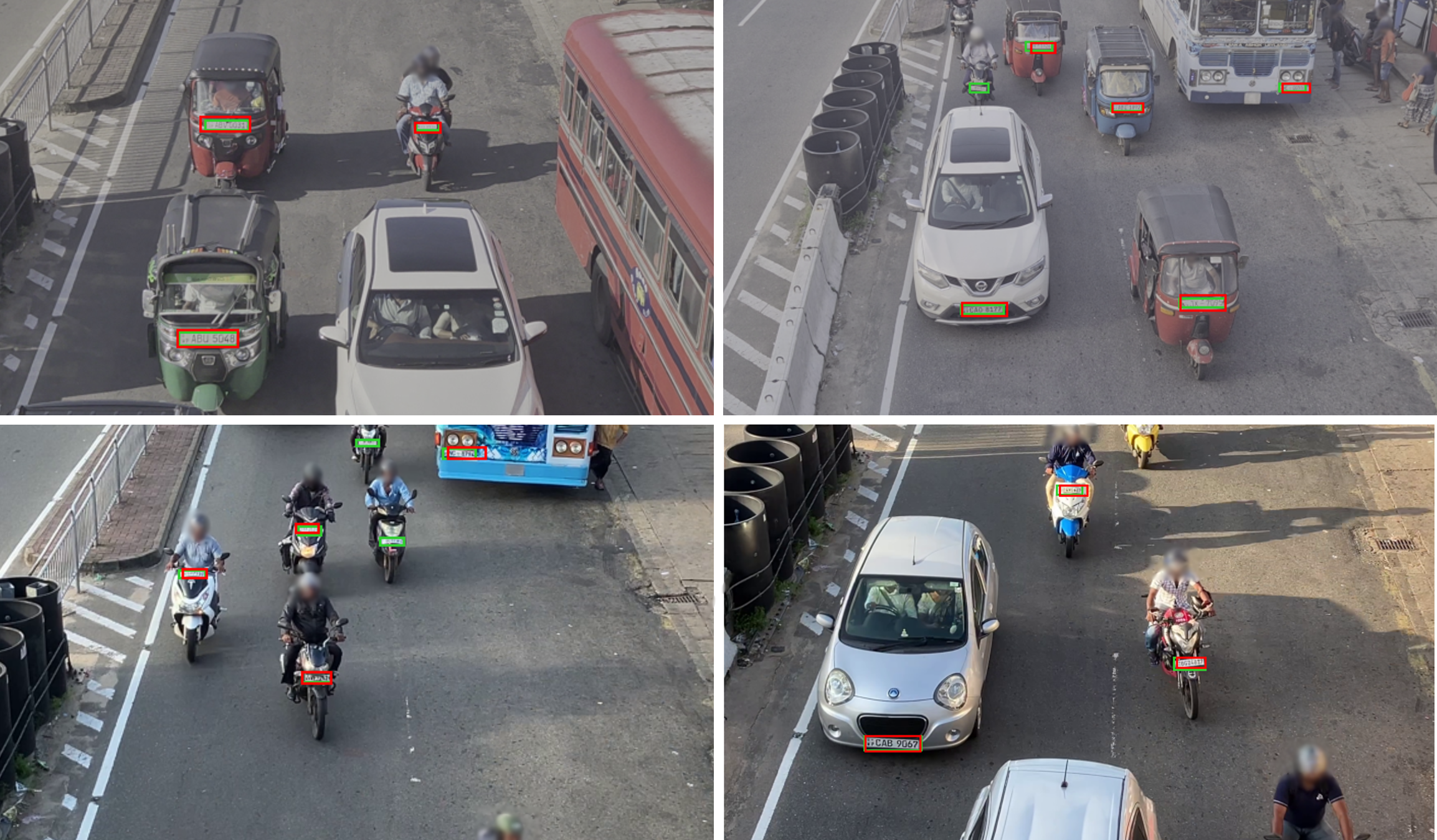}
    \caption{{Predicted (red) and ground truth (green) bounding boxes on test set images in the SL-LPR dataset.}}
    \label{fig:testresults}
\end{figure}

Training the model specifically on a diverse multi-vehicle dataset has significantly improved its performance. Figure~\ref{fig:oldnewresult} shows the predicted license plates in a complex traffic scene with (a) the model trained only on the PKU dataset~\cite{chinesedataset} and (b) after fine-tuning on the SL-LPR dataset. Since the PKU dataset mostly contains cars, the initial model has failed to detect license plates of the visually different motorbikes and buses. After fine-tuning the model on the SL-LPR dataset, it has successfully detected license plates on all vehicle types. Furthermore, there are no detections of other text on the bus, indicating that the model can distinguish license plates from unrelated text. These results show that our LPD model can handle complex unstructured traffic scenarios accurately. %

\begin{figure}[htbp]
    \centering
    \begin{subfigure}{0.48\linewidth}
        \centering
        \includegraphics[width=\linewidth]{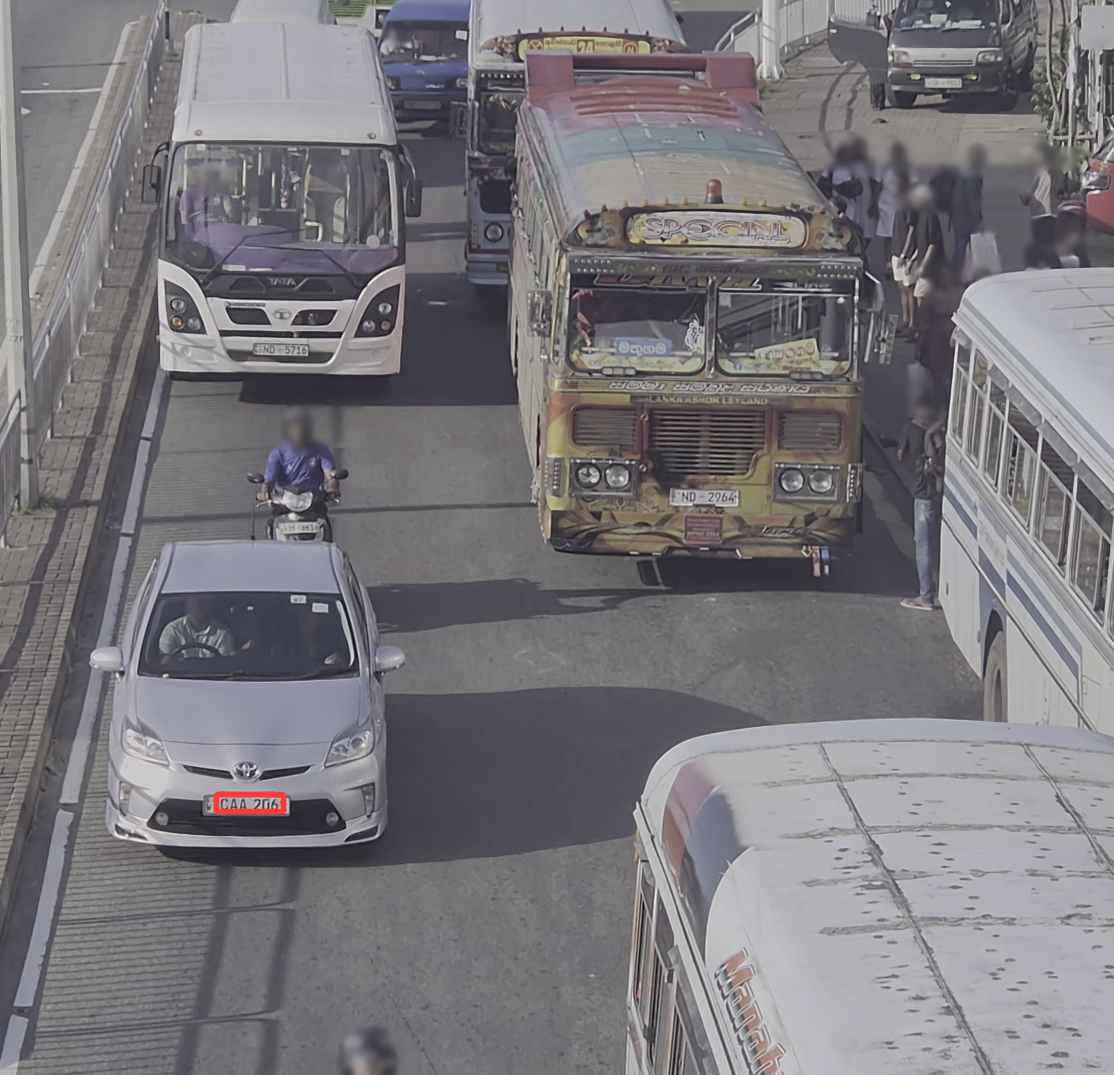}
        \caption{Detections from model trained on the PKU dataset~\cite{chinesedataset}}
    \end{subfigure}
    \hfill
    \begin{subfigure}{0.48\linewidth}
        \centering
        \includegraphics[width=\linewidth]{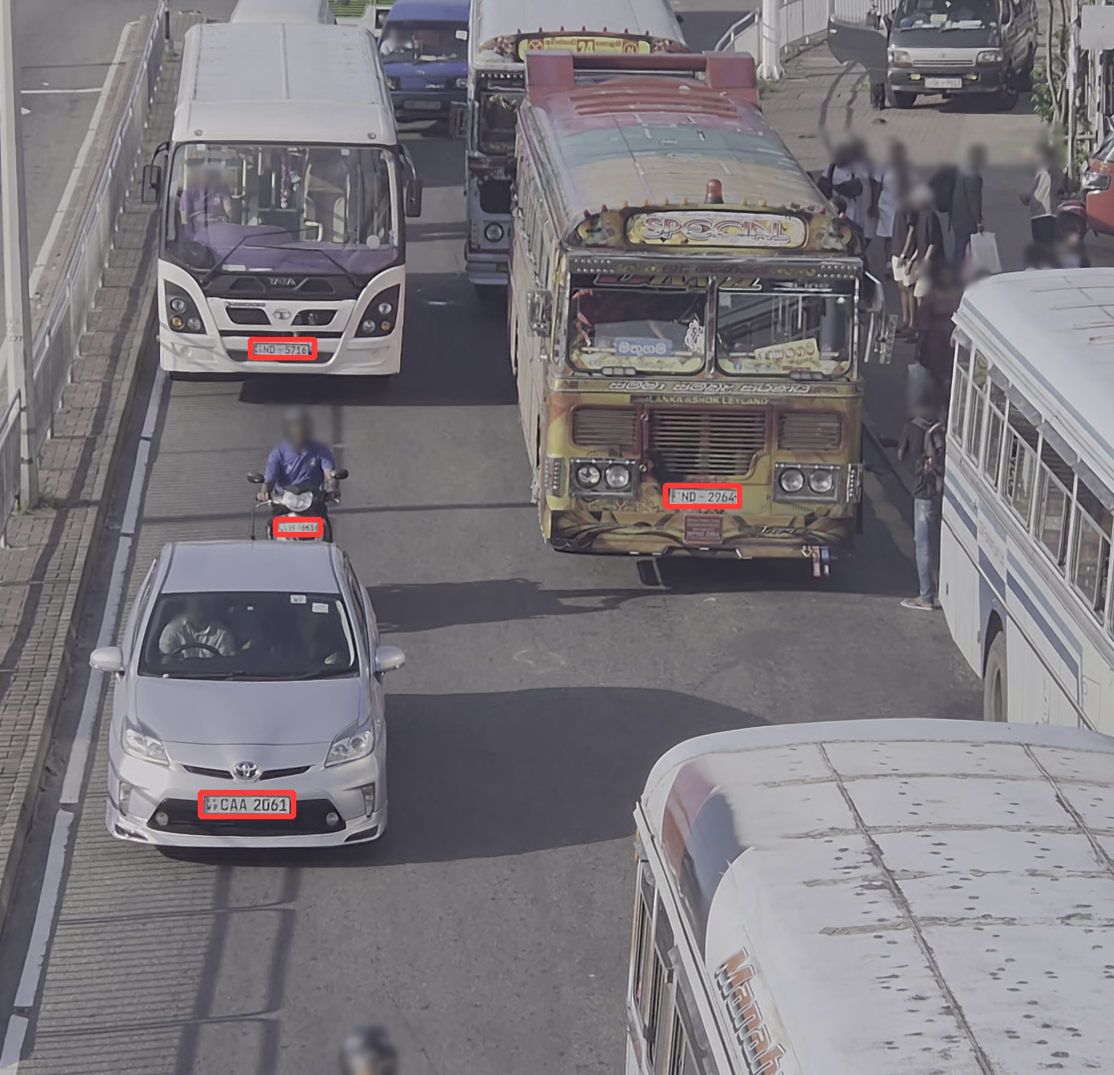}
        \caption{Detections from model fine-tuned on the SL-LPR dataset}
    \end{subfigure}
    \caption{Comparison of detected license plates before and after fine-tuning the model on the SL-LPR dataset. (b) is more effective for detection on multiple vehicle types.}
    \label{fig:oldnewresult}
\end{figure}


\subsection{License Plate Character Recognition}

We trained the character recognition model for 700 epochs using an initial learning rate of 0.001, which was decayed by 15\% if the loss did not improve for 30 consecutive epochs. On the SL-LPR dataset, the model achieved a character-level accuracy of 97.42\% and a plate-level accuracy of 87.88\%. 



In Table~\ref{tab:performance}, we compare our LPCR model with prior work evaluated on the PKU~\cite{chinesedataset} and AOLP~\cite{hsu2012application} datasets, and the Argentinian license plate dataset (Arg-plate) found in the fast-plate-ocr repository~\cite{fastplateocr}. Our model achieves a comparable accuracy to the fast-plate-ocr model on both the Arg-plate and SL-LPR datasets, while being smaller and quantized to suit embedded implementation. The lower plate-level accuracy on the SL-LPR dataset can be attributed to the dataset having more unclear images compared to the other datasets. However, our model achieves a high character-level accuracy in all cases, indicating that the plate-level errors occurred from only a few mispredicted characters. 

\begin{table}[htbp]
\centering
\caption{Comparison of our LPCR model with other implementations based on plate-level accuracy (\%) on several public datasets and latency as reported in the original work.}
\label{tab:performance}
\begin{tabular}{lC{0.6cm}C{0.6cm}C{0.6cm}C{0.6cm}C{1.3cm}}
\toprule
\textbf{Method} & \textbf{PKU} & \textbf{AOLP} & \textbf{Arg-plate} & \textbf{SL-LPR} & \textbf{Latency (ms)} \\
\midrule
Li et al.~\cite{li2024mixed} & \textbf{99.5} & -- & -- & -- & 50 [FPGA]\\
Zhang et al.~\cite{zhang2020robust}      & 88.2  & 95.8 & -- & -- & 25 [GPU] \\
Hsu et al.~\cite{hsu2012application}  & --  & 91.2 & -- & -- & 90 [CPU] \\
fast-plate-ocr~\cite{fastplateocr} & --  & -- & 94.1 & \textbf{90.0} & 13 [CPU] \\
\midrule
\textbf{Ours}                            & 94.2 & \textbf{96.2} & \textbf{95.3} & 87.9 & \textbf{4 [FPGA]} \\
\bottomrule
\end{tabular}
\end{table}

{

Our model surpasses several prior methods on the PKU~\cite{chinesedataset} and AOLP~\cite{hsu2012application} datasets, validating its adaptability to other alphabets such as Chinese. Furthermore, the work of Li et al.~\cite{li2024mixed} is directly comparable to our method as it is also implemented on the Kria KV260 platform. While it has higher accuracy due to its transformer-based approach, it is about 12$\times$ slower per license plate, which could make it impractical for real-time use in busy multi-vehicle scenarios. 

}
\begin{figure}[htbp]
    \centering
    \includegraphics[width=0.49\textwidth]{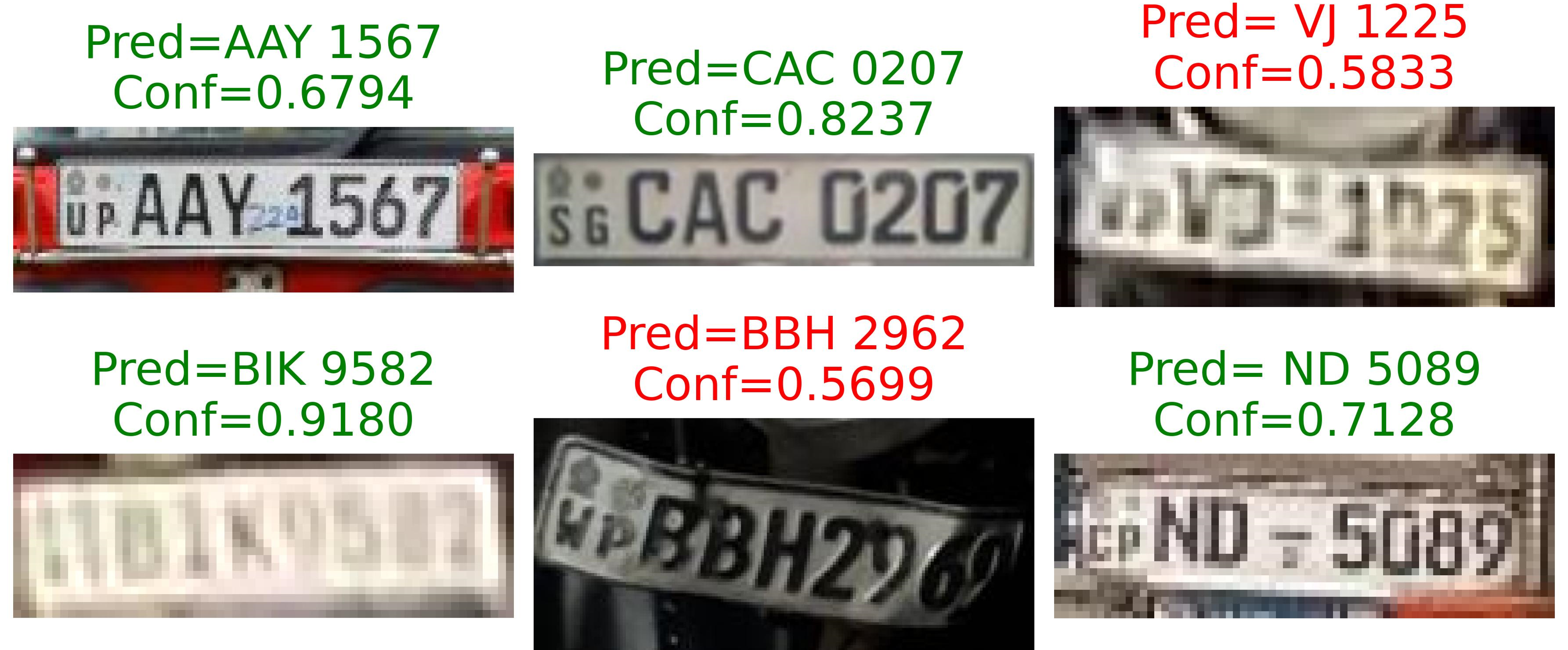}
    \caption{{Predictions of our model for Sri Lankan license plates of different types and conditions.}}
    \label{fig:ourmodel}
\end{figure}

{
Figure~\ref{fig:ourmodel} shows that our model is capable of accurately recognizing license plate characters, while ignoring the two small characters at the start, even when the cropped image shows part of the background. It can also recognize license plates that are slightly blurred but readable. However, in cases where the plate is partially obscured, the model may produce minor errors, typically involving only a single character. These errors are identifiable from the lower confidence levels compared to correctly recognized plates. In the end-to-end system, to prevent the application from recording incorrect license plate numbers, we replace the predicted character with a space if its confidence is lower than a selected threshold.}

{
In Figure~\ref{fig:both}, we show the combined performance of license plate detection and character recognition by cropping the bounding boxes predicted by the detection model and feeding them into the character recognition model. Almost all license plates have been recognized accurately, even when the bounding box is not exactly the edge of the plate. Furthermore, both models are lightweight, having only 1.4~million parameters combined, with a parameter storage size of 268~KB after quantization. Therefore, these two models can be successfully combined to create an efficient end-to-end license plate recognition system.
}

\begin{figure}[h!]
    \centering
    \includegraphics[width=0.49\textwidth]{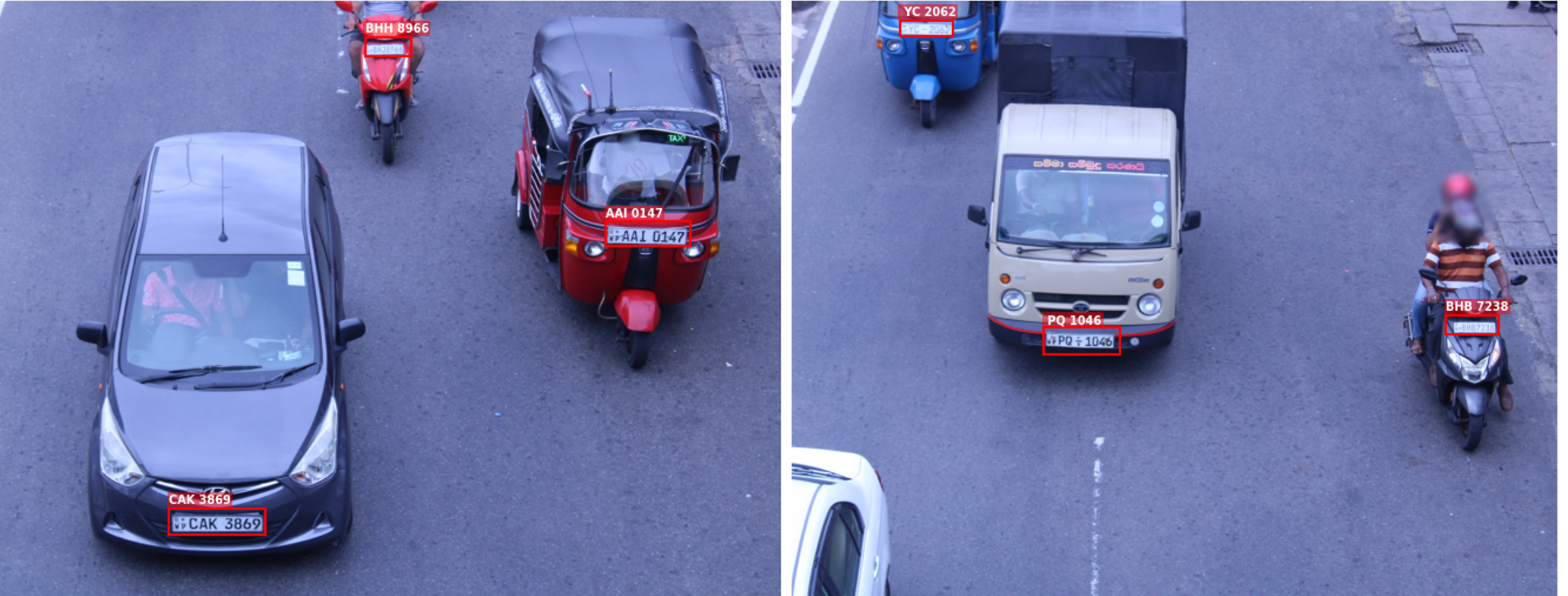}
    \caption{{Combined results of license plate detection and recognition using the developed models.}}
    \label{fig:both}
\end{figure}

\subsection{Hardware Implementation}

We verified that the accuracy figures from the above sections were maintained throughout the various steps of the hardware implementation process, both in simulations and in the final FPGA deployment. Furthermore, we integrated a camera into the system and observed the accuracy of the license plate recognition process in real-time field tests. The end-to-end system has a power consumption of 4.2~W, showing improved power efficiency over GPU-based approaches. 


\begin{table}[htbp]
\centering
\caption{Latency of each part of our system, compared with the latency when run on CPU-only systems. Server-grade CPU: Intel Xeon Silver 4210 (10 cores, 2.20 GHz); embedded CPU: ARM Cortex-A53 (4 cores, 1.3 GHz). Our system, implemented on the Kria KV260, divides functions between its FPGA fabric and processing system (PS).}
\label{tab:latency}
\begin{tabular}{m{3cm}C{1.5cm}C{1.2cm}C{1.2cm}}
\toprule
\textbf{Function} & \textbf{Our system (ms)} & \textbf{Server CPU (ms)} & \textbf{Embedded CPU (ms)} \\
\midrule
LPD (per frame) & \textbf{9.9} [FPGA] & 9.1 & 128.2 \\
LPCR (per plate) & \textbf{4.0} [FPGA] & 4.7 & 39.5 \\
Frame resizing & 18.1 [PS] & 9.7 & 18.1\\
Camera image acquisition & 6.0 [PS] & 6.0 & 6.0\\
Other intermediate steps & 10.0 [PS] & 4.5 & 10.0\\
\midrule
\textbf{Total time for a frame with 10 license plates} & \textbf{87.0} & 76.3 & 557.3\\
\bottomrule
\end{tabular}
\end{table}

Table~\ref{tab:latency} shows the latency of different components of our system, compared to the same functions executed on an Intel Xeon server-grade CPU and an ARM Cortex-A53 embedded processor. Since the Kria KV260 contains the same ARM processor, the non-accelerated steps have the same latency in our system and the embedded CPU-only system. In the accelerated steps, our system achieves comparable performance to the server-grade CPU while vastly outperforming the embedded CPU implementation. Currently, the frame resizing step is the bottleneck of the system, as it involves a high-resolution image and is done on the PS. However, its impact on the throughput can be reduced by using a multi-threaded implementation. Our system is capable of processing a frame containing up to 10 license plates within a maximum time of 87 ms, hence operating at an 11.5~FPS throughput in most urban road scenarios. Therefore, as mentioned in Section~III, it can be used with a 10 FPS camera for real-time LPR on vehicles traveling at speeds of up to 250 km/h.

It is clear that using FPGA acceleration has significantly improved the model inference latencies. Moreover, using an FPGA instead of a GPU to achieve this acceleration reduces the power consumption and production cost. Therefore, this system is a promising solution for high-performing automated traffic monitoring systems in developing countries.
    
	\section{Conclusion and Future Work}
	\label{sec:conclusion}
In this paper, we presented an embedded real-time license plate recognition system customized for challenging road scenarios seen in developing countries. Our approach comprises two FPGA-accelerated models for license plate detection and license plate character recognition. We adapted a low-precision YOLO model~\cite{gunay2022lpyolo} for license plate detection and optimized the fast-plate-ocr CNN architecture~\cite{fastplateocr} for character recognition. We also created a robust dataset of Sri Lankan road and license plate images for training and testing. Through this dataset and models, we developed a method that can be successfully used even in high-traffic complex scenes with any type of vehicle.

We addressed the need for real-time operation in an embedded environment by training quantized models using Brevitas~\cite{brevitas} and compiling them into FPGA designs using the FINN framework~\cite{finn}.  With FPGA acceleration, we achieved faster processing than a CPU-only implementation while requiring less power and cost compared to a GPU. The complete license plate recognition pipeline was implemented on the Xilinx Kria KV260 platform with camera integration for real-time usage at 11.5 FPS.

{This LPR system demonstrates strong potential to be used in intelligent transportation systems in developing countries. As future work, the SL-LPR dataset can be expanded to include nighttime images and more location variety to ensure accurate detection in all environments. Future research may also explore integration with other technologies to enable various applications, such as combining LPR with speed estimation for speed enforcement or traffic analysis. }

\section*{Acknowledgements}
	The authors thank the National Research Council of Sri Lanka (Grant No. 19-080) for computational resources, and the Sri Lanka Police, the Road Development Authority, and the National Council for Road Safety for their support.

	
	\bibliographystyle{IEEEtran}
	\bibliography{root} 
	
\end{document}